\documentclass[journal]{IEEEtran}
\usepackage{microtype}
\usepackage{graphicx}
\usepackage{amsmath}
\usepackage{todonotes}
\usepackage{graphicx}
\usepackage{calc}
\usepackage{subcaption}
\usepackage{pdfpages}
\usepackage{url}

\ifCLASSINFOpdf
\else
\fi

\begin{document}

\title{Detecting Blinks in Healthy and Parkinson's EEG\\ \LARGE{A Deep Learning Perspective}}

\author{
Artem Lensky\IEEEauthorrefmark{1}\IEEEauthorrefmark{2},
Yiding Qiu\IEEEauthorrefmark{1}\\
\IEEEauthorblockA{\IEEEauthorrefmark{1} School of Engineering and Technology, University of New South Wales, Canberra ACT, Australia}\\
\IEEEauthorblockA{\IEEEauthorrefmark{2} School of Biomedical Engineering,
University of Sydney, NSW, Sydney, Australia}\\
\thanks{Both authors contributed equally to this research.}
\thanks{Corresponding author: Artem Lensky (email: a.lenskiy@unsw.edu.au).}}


\maketitle

\begin{abstract}
\textit{Introduction:} Blinks in electroencephalography (EEG) are often treated as unwanted artifacts. However, recent studies have demonstrated that blink rate and its variability are important physiological markers to monitor cognitive load, attention, and potential neurological disorders. This paper aims to address the critical task of accurate blink detection by evaluating various deep learning models for segmenting EEG signals into involuntary blinks and non-blinks. 

\textit{Methods:} The paper presents a pipeline for blink detection using 1, 3, or 5 frontal EEG electrodes. The problem is formulated as the sequence-to-sequence problem and tested on various deep learning architectures including standard recurrent neural networks, convolutional neural networks (both standard and depth-wise), temporal convolutional networks (TCN), transformer-based models, and hybrid architectures. The models were trained on raw EEG signals with minimal pre-processing. Training and testing was carried out on a public dataset of 31 subjects collected at UCSD. This dataset consisted of 15 healthy participants and 16 patients with Parkinson's disease allowing us to verify the model's robustness to a tremor.

\textit{Results:} Out of all models, CNN-RNN hybrid model consistently outperformed other models and achieved the best blink detection accuracy of 93. 8\%, 95.4 and 95. 8\% with 1, 3, and 5 channels in the health cohort and correspondingly 73. 8\%, 75. 4\% and 75. 8\% in patients with PD.
\textit{Conclusion:} The paper compares neural networks for the task of segmenting EEG recordings to involuntarily blinks and no blinks allowing for computing blink rate and other statistics.
\end{abstract}

\section{Introduction}
Electroencephalography (EEG) records electrical activity over the human scalp, capturing signals generated by brain activity as well as artifacts. Artifacts in EEG signals are broadly categorized as physiological and non-physiological. Non-physiological artifacts arise from external sources, such as electrical interference or mechanical impact (e.g., touching an EEG cap). Physiological artifacts, on the other hand, originate from muscle activity, including cardiovascular functions, body movement, and facial muscle contractions, which are responsible for ocular activity like blinking.

Blinking involves two muscle groups: the levator palpebrae superioris and orbicularis oculi muscles, which are skeletal muscles, and the superior and inferior tarsal muscles, which are smooth muscles. These muscles act antagonistically, and their contractions produce high-amplitude peaks in EEG signals, often exceeding 100$\mu V$.

Blinks can be classified into three types:
\begin{itemize}

\item[] - Reflex blinks, triggered by external stimuli like light flashes or loud sounds, are frequently used to study the interaction between the central nervous system and the neuronal blink circuit.

\item[] - Voluntary blinks, initiated by the cerebral cortex, occur in response to conscious effort.

\item[] - Spontaneous blinks, which occur without identifiable stimuli, serve essential functions such as maintaining eye lubrication but are challenging to study due to their lack of an external trigger.

\end{itemize}

Accurate detection of blink events in EEG is essential for understanding their physiological implications. This includes pinpointing the exact onset and offset of each blink. While alternative methods like ocular electromyography or eye-facing cameras offer precise detection, they require additional hardware and experimental design. In contrast, EEG-based detection leverages the widespread availability of experimental EEG data, providing a cost-effective approach to analyze blinks without extra experimental overhead.

Blink artifacts predominantly affect frontal electrodes \texttt{Fp1}, \texttt{Fp2}, \texttt{Fz}, \texttt{F3}, and \texttt{F4}, located on the forehead. The proximity of these electrodes to the eyes makes them highly sensitive to blink-induced muscle contractions, which manifest as sharp, high-amplitude peaks. Identifying these artifacts accurately enables the extraction of blink rate and variability, physiological markers that have been shown to correlate with attention and cognitive performance \cite{yajima2018development,paprocki2017}. Furthermore, altered blink patterns have been observed in neurological conditions such as Parkinson's disease (PD), where reduced blink rates and tremor-induced artifacts pose unique challenges to analysis \cite{delorme2004eeglab}.

Recent advances in deep learning have provided robust tools for analyzing complex temporal patterns in EEG data. Deep learning models, such as convolutional neural networks (CNNs) and recurrent neural networks (RNNs), excel at capturing spatial features (e.g., sharp peaks caused by blinks) and temporal dependencies (e.g., sequential blink patterns). These models facilitate automated and accurate blink detection, enabling not only artifact removal but also the identification of clinically relevant biomarkers. This study evaluates various deep learning architectures for their effectiveness in detecting blinks across both healthy individuals and PD patients, aiming to fill a gap in the literature by providing a systematic comparison of model performance and robustness.

\section{Related Work}

With the proliferation of wearable devices and headsets, the task of detecting blinks from electroencephalograms (EEG) or electrooculograms (EOG) has garnered considerable attention. This section reviews methods for blink detection in EEG, categorized into two main groups: non-deep-learning approaches and deep-learning-based approaches.

\subsection{Non-deep-learning approaches to blink detection}

Before deep learning became prevalent in blink detection, various approaches relied on hand-crafted features and classical machine learning models. For instance, Kim et al.~\cite{kim2007automatic} focused on automatic spike detection in neural activity, laying the foundation for many blink detection algorithms. Their method employed template matching to iteratively estimate the morphology of action potentials. Specifically, the process involved estimating signal power, applying a low-pass filter, identifying extreme modes, and refining templates to isolate distinct patterns from noise.

Paprocki et al.~\cite{paprocki2016} proposed a blink localization method using independent component analysis (ICA) on preprocessed \texttt{Fp1-F3} and \texttt{Fp2-F4} EEG channels. Their pipeline removed outlier amplitudes and employed band-pass filtering and normalization. Blink candidates were identified based on morphological features, including the steepness of the waveform's ascending and descending parts. This approach demonstrated an effective way to handle artifacts while preserving blink-related components.

Kleifges et al. developed BLINKER~\cite{kleifges2017blinker}, a comprehensive pipeline for detecting blinks and extracting metrics such as blink rate, duration, and velocity-amplitude ratios. The method relied on bandpass filtering to enhance blink signals and employed thresholds to identify candidates exceeding 1.5 standard deviations from the mean. Morphological features were extracted to further refine the detection. The pipeline's implementation as a MATLAB toolbox facilitated its adoption in large-scale studies, showcasing its utility.

Jalilifard et al.~\cite{Jalilifard2020} extended BLINKER's framework to single-channel EEG. They extracted 26 features and trained classifiers, achieving a person identification accuracy of 98.7\% using blink waveforms. This study illustrated the feasibility of using EOG-derived features for human authentication.

Agarwal et al.~\cite{agarwal2019blink} introduced a single-channel blink detection algorithm that applied signal filtering and a peak detection method. This simple yet effective approach provided a baseline for further refinement in blink identification.

Cao et al.~\cite{cao2021unsupervised} proposed an unsupervised algorithm leveraging cross-channel correlations between \texttt{Fp1} and \texttt{Fp2}. The method also considered amplitude displacement distributions and Gaussian mixture models for blink detection. This unsupervised framework demonstrated potential for robust artifact handling without requiring labeled data.

Wang et al.~\cite{wang2022multidimensional} developed a hybrid approach combining nonlinear energy operators and K-means clustering. Their method selected discriminative features through variance filtering and incorporated support vector machines for final artifact classification.

Zhang et al.~\cite{Zhang2023} proposed RT-Blink, a real-time blink detection method employing a random forest classifier. The system used wavelet transforms and threshold algorithms to achieve efficient and accurate blink detection, making it suitable for applications requiring low computational overhead.

\subsection{Deep learning approaches to blink detection}

The introduction of AlexNet in 2012 revolutionized the adoption of convolutional neural networks (CNNs) across domains. Deep learning models, including recurrent neural networks (RNNs) and transformers, have since gained prominence in ocular artifact detection.

Lo Giudice et al.~\cite{logiudice2020convolutional} introduced a one-dimensional CNN to differentiate between voluntary and involuntary blinks. Their model consisted of a convolutional layer, dropout, and pooling layers. This simple architecture was effective for distinguishing blink types and found applications in brain-computer interfacing.

Sawangjai et al.~\cite{sawangjai2022eeganet} presented EEGANet, a generative adversarial network (GAN) designed for artifact cleaning. The generator learned to remove artifacts, while the discriminator validated whether the signals were real or synthetic. This innovative approach tackled both artifact cleaning and classification simultaneously.

Reyes et al.~\cite{reyes2021lstm} applied long short-term memory (LSTM) networks to detect spontaneous blinks and map them to character inputs. Their method demonstrated utility in assistive technologies, enabling communication through blink patterns.

Jurczak et al.~\cite{jurczak2022cnn} proposed a convolutional neural network (CNN) trained to remove eye blink artifacts from EEG signals. This method demonstrated superior performance over traditional techniques like Independent Component Analysis (ICA), particularly for central electrodes.

Iaquinta et al.~\cite{iaquinta2021eeg} introduced an automated blink detection system using CNNs. Their algorithm achieved high reliability across multiple public EEG datasets, demonstrating robustness in diverse conditions.

Sun et al.~\cite{sun2020rescnn} presented a one-dimensional residual convolutional neural network (1D-ResCNN) for EEG denoising, effectively mapping noisy signals to clean signals, and showcasing its applicability for blink artifact removal.

In summary, while traditional and deep learning approaches have advanced blink detection significantly, challenges such as segmenting the opening and closing phases of blinks remain. Future research should explore hybrid models and advanced architectures to address these limitations.

\section{Methods}

Several publicly available EEG datasets were considered for training models and evaluating their performance in blink detection and segmentation tasks. However, many were excluded during the initial screening due to the following reasons: (a) datasets with preprocessed EEG recordings where blink artifacts were already removed, (b) datasets designed for experiments involving subjects with eyes closed, and (c) datasets lacking both healthy controls and patients with neurological disorders, which are crucial for testing model robustness. Among the screened datasets, the UC San Diego dataset met all the criteria. This dataset includes EEG recordings from 31 subjects: 15 healthy controls and 16 Parkinson's disease (PD) patients. PD patients often exhibit resting tremors, which present additional challenges for blink detection due to signal contamination. The EEG recordings were sampled at 512 Hz.

\begin{figure*}[ht]
\centering
  \includegraphics[trim={0.3cm 7.3cm 0.3cm 6.0cm},clip,width=1\linewidth]{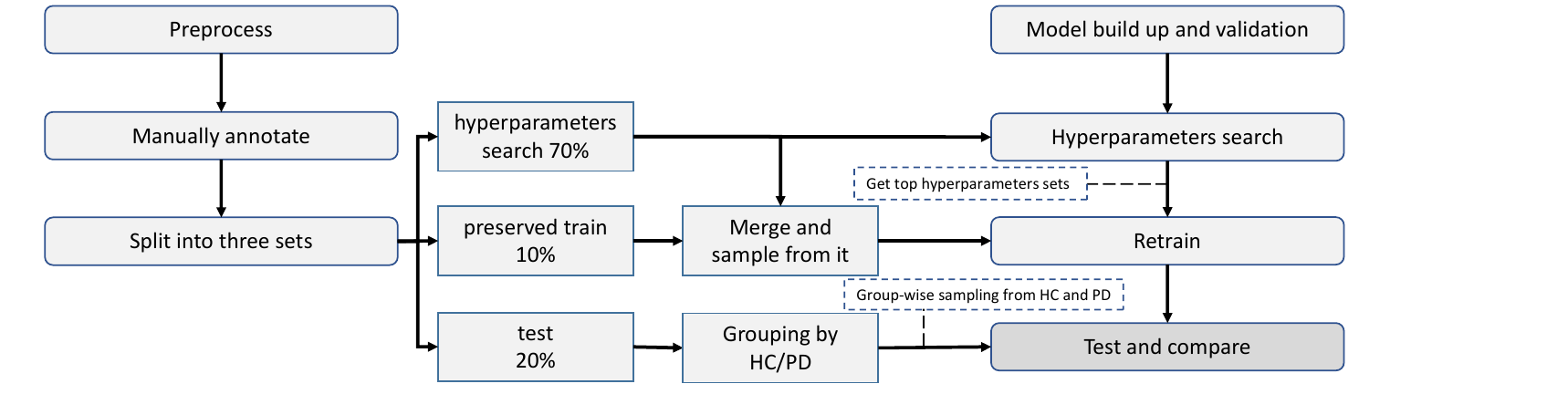}
  \caption{The experimental pipeline consists of two parts: the hyperparameter search and the validation of the models on the best sets of hyperparameters. Validation was performed separately on EEG data from healthy controls (3 subjects) and PD patients (3 subjects), collectively accounting for 20\% of the dataset.
  \label{fig:pipeline}}
\end{figure*}

The training, hyperparameter search, and evaluation processes were performed using an RTX A5000 GPU, following the pipeline depicted in Fig.~\ref{fig:pipeline}. Hyperparameter tuning was essential for optimizing model performance and ensuring fair comparisons across architectures.

\subsection{Selected Electrodes}
\begin{figure}[ht]
    \includegraphics[trim={6.5cm 10.18cm 7.5cm 0.1cm},clip,width=\linewidth]{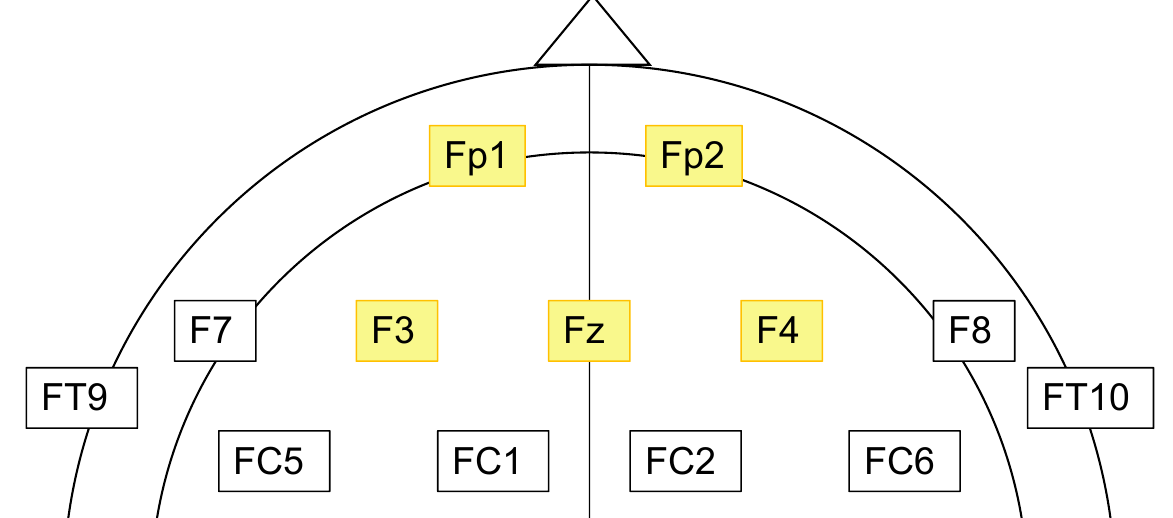}
    \caption{Forehead part of the standard 10--20 system EEG electrode placement. The nose is denoted by a triangle, defining the upward facial direction. Front electrodes \texttt{Fp1}, \texttt{Fp2}, \texttt{Fz}, \texttt{F3}, and \texttt{F4} are highlighted in yellow.}
    \label{electrodsPlacment}
\end{figure}

The following five frontal electrodes, shown in Fig.~\ref{electrodsPlacment}, were used in this study: \texttt{Fp1}, \texttt{Fp2}, \texttt{Fz}, \texttt{F3}, and \texttt{F4}. These electrodes were chosen due to their proximity to the eyes, making them highly sensitive to blink-induced artifacts. \texttt{Fp1} and \texttt{Fp2}, located directly above the left and right eyes, respectively, are the most affected by blink artifacts, while \texttt{Fz}, located centrally on the forehead, exhibits slightly reduced blink amplitudes \cite{Nakanishi2012}. The relative contributions of \texttt{F3} and \texttt{F4}, situated laterally above \texttt{Fp1} and \texttt{Fp2}, were also investigated.

Experiments were conducted using three electrode configurations: single-channel (e.g., \texttt{Fp1}), three-channel (\texttt{Fp1}, \texttt{Fz}, \texttt{Fp2}), and five-channel (all selected electrodes). Single-channel configurations simulate wearable devices with limited electrodes, while multi-channel setups test the added benefit of spatial information for improving blink segmentation accuracy.

\subsection{Problem Formulation}
Blink detection was formulated as a sequence-to-sequence task. The input consisted of one- to five-dimensional temporal signals from the selected electrodes, while the output was a one-dimensional label sequence indicating whether each time point corresponded to a blink or not. 

Some models, such as those with CNN blocks, require fixed window sizes for processing due to their architectural constraints. Consequently, blink events occurring near the boundaries of windows might be truncated, leading to incomplete detection. To address this limitation, a weighted voting algorithm was implemented in the post-processing stage. This algorithm involves shifting the sampling windows by multiple offsets and aggregating predictions across overlapping segments. For each time point, the final label is determined by majority voting, significantly improving edge-case accuracy.

\begin{figure}[ht]
    \centering
      \includegraphics[trim={3.0cm 11cm 2.5cm 11cm},clip,width=\linewidth]{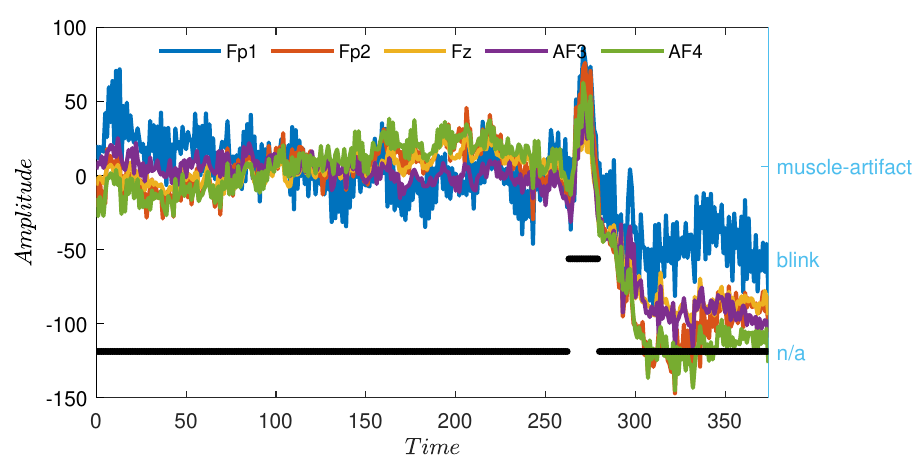}
      \caption{Illustration of signals from the five selected electrodes. The black line represents the ground truth labels for blinks.}
      \label{fig:}
\end{figure}

\subsection{Tested Models}
Several deep learning models were explored, broadly categorized into recurrent neural networks (RNNs), convolutional neural networks (CNNs), and hybrid models. Each model was chosen for its demonstrated effectiveness in time-series analysis, with the following specific characteristics:

\textbf{RNNs}: This category includes Long Short-Term Memory (LSTM), Bidirectional LSTM (biLSTM), Gated Recurrent Unit (GRU), and Bidirectional GRU (biGRU). RNNs excel at capturing temporal dependencies in sequential data. Bidirectional variants, such as biLSTM and biGRU, process data in both forward and backward directions, enhancing the extraction of temporal patterns \cite{Rahman2022}.

\textbf{CNNs}: Standard CNNs and Temporal Convolutional Networks (TCNs) were utilized. TCNs employ dilated causal convolutions, allowing them to capture long-term dependencies while maintaining computational efficiency \cite{TCN}. Depth-wise CNNs, inspired by MobileNet \cite{Howard2017}, were also tested to reduce model complexity while maintaining performance. These models are particularly suited for real-time applications where computational resources may be constrained.

\textbf{Hybrid Models}: Hybrid architectures, such as CNN-RNN and TCN-RNN, combine the feature extraction capabilities of CNNs with the sequential modeling power of RNNs. Depth-wise variations of these hybrids were also implemented, offering a balance between accuracy and computational efficiency. Hybrid models are well-suited for blink detection tasks, where spatial features (e.g., waveform shapes) and temporal patterns (e.g., blink intervals) both play critical roles.

The performance of each model was rigorously evaluated using grid-based hyperparameter search to identify the optimal architecture for accurate and robust blink detection.

\section{Results}
    \subsection{Performance metrics}
    In evaluating the performance of the aforementioned models, we compute the following metrics: F1-Score-micro, F1-Score-macro, recall rate at the level of the entire blink signal , and precision rate at the level of the entire blink signal.
    The F1 score-micro is defined as the evaluation of the model's performance at each individual time point. In contrast, the F1-Score-macro treats each complete blink signal as a single example for evaluation purposes.
        
    However, given that blinks actually occupy only a small fraction of the time, F1 scores were chosen as the main performance metric. Accuracy, on the other hand, can be misleading in our context, as a model that simply classifies everything as ``N/A" (as a label for no blink) would produce a high accuracy and hence would not truly reflect the models' performance. The F1 score (Eq. \ref{eg:1}) considers both precision and recall, and provides a more balanced measure by penalizing false positives and negatives. It is preferable in situations with class imbalance data, as it considers both false positives and false negatives. 
        \begin{equation} 
        \text{F1-Score} = 2 \cdot \frac{\text{Precision} \cdot \text{Recall}}{\text{Precision} + \text{Recall}}
        \label{eg:1}
        \end{equation}
    where:
        \begin{equation*}
        \text{Precision} = \frac{TP}{TP + FP}, \quad \text{Recall} = \frac{TP}{TP + FN}
        \end{equation*}
        
    where, TP (True Positives) represents the number of correctly identified positive instances, FP (False Positives) is the number of instances incorrectly labeled as positive, and FN (False Negatives) is the number of positive instances that were incorrectly labeled as negative. The F1 score balances precision and recall, providing a metric that reflects the model's performance in handling false positives and false negatives. The F1 scores may yield lower values, but better capture the model's performance.

    For example, if the model assigns ``N/A" to all samples, with an accuracy of 72\% due to ``N/A" comprising 72\% of the labels, achieve a far better accuracy than a random guess. However, this accuracy can not represent the performance of the model. In this analysis, the F1 score is calculated as the harmonic mean of precision and recall, as shown in Equation \ref{eg:1}, the all ``N/A'' result will have a score of 0. If the model does random guess, then for the ``blinks'' class, the precision, which represents the proportion of correctly identified ``blinks'' among all predicted ``blinks'', is calculated to be 0.28. The recall, indicating the proportion of actual 'blinks' that were correctly identified, is 0.50. These values lead to an F1 score of 0.3590 for the ``blinks'' class. The overall F1-Score, also considering the F1-Score of ``N/A'', as 0.72, is around 0.4745. These indicate that the F1 scores of the meaningful models' predictions should be significantly higher than these baseline values.

    \subsection{Hyperparameter search}
    
    A grid based hyperparameter search was conducted over a different set of hyperparameter combinations on all studied models. The heatmap visualisation is provided at the end of the paper, to visually compare the relationships between the models and the underlying parameters. 
    During the comparison process, we ensured the consistency of parameter combination selections as much as possible. However, it is important to note that depending on the model the impact of the same hyperparamter is different. For example, adding a neuron in CNN and RNN results in a different number of additional weights.
    
    
    \begin{table*}[t]
        \centering
        \caption{\label{table:hyperparameters}Comparison of hyperparameters selection during the search}
        \label{tab:model_comparison}
        \begin{tabular}{|l|c|c|c|c|c|c|}
        \hline
        \textbf{Model} & \textbf{Filter Size} & \textbf{Num Blocks} & \textbf{Num Channels} & \textbf{Num Filters} & \textbf{Num RNN Blocks} & \textbf{Num Units} \\
        \hline
        CNN-DW & 5 11 15 & 1 2 3 4 & 1 3 5 & 8 16 32 & - & - \\
        CNN-ST & 5 11 15 & 1 2 3 4 & 1 3 5 & 8 16 32 & - & - \\
        CNN-RNN-DW & 5 15 & 1 2 3 & 1 3 5 & 8 16 32 & 1 2 & 8 16 32 \\
        CNN-RNN-ST & 5 15 & 1 2 3 & 1 3 5 & 8 16 32 & 1 2 & 8 16 32 \\
        RNN (bilstmLayer) & - & - & 1 3 5 & - & 1 2 3 4 & 8 16 32 \\
        RNN (gruLayer) & - & - & 1 3 5 & - & 1 2 3 4 & 8 16 32 \\
        RNN (lstmLayer) & - & - & 1 3 5 & - & 1 2 3 4 & 8 16 32 \\
        TCN-DW & 5 11 15 & 1 2 3 4 & 1 3 5 & 8 16 32 & - & - \\
        TCN-ST & 5 11 15 & 1 2 3 4 & 1 3 5 & 8 16 32 & - & - \\
        TCN-RNN & 5 11 15 & 1 2 3 & 1 3 5 & 8 16 32 & 1 2 & 8 16 32 \\
        TCN-RNN-ST & 5 11 15 & 1 2 3 & 1 3 5 & 8 16 32 & 1 2 & 8 16 32 \\
        \hline
        \end{tabular}\\
        This table presents various neural network models and their corresponding hyperparameters. A dash ``-" indicates that the parameter is not applicable for that particular model. 
    \end{table*}
    
    A grid hyperparameter search was chosen as the most suitable for deducing which of the parameters impact the model performance the most.  Grid search also allowed us to observe that some models are more parameter-sensitive than others, to the point that some of the models performance often was below the baseline. While other models were less sensitive to variations in parameters. One of such models is CNN-RNN.

    
    \begin{table*}[t]
        \centering
        \caption{\label{table:best}Best F1-Score-micro for each model and corresponding hyperparameters}
        \label{tab:model_comparison}
        \begin{tabular}{|l|c|c|c|c|c|c|}
        \hline
        \textbf{Model} & \textbf{Filter Size} & \textbf{Num Blocks} & \textbf{Num Filters} & \textbf{Num RNN Blocks} & \textbf{Num Units} & \textbf{Mean F1 Score} \\
        \hline
        CNN-RNN & 15 & 2 & 32 & 2 & 32 & 0.93178 \\
        TCN-RNN & 5 & 3 & 32 & 1 & 32 & 0.92638 \\
        CNN-ST & 15 & 4 & 32 & - & - & 0.91132 \\
        CNN-DW & 15 & 4 & 32 & - & - & 0.90963 \\
        TCN-DW & 15 & 4 & 32 & - & - & 0.90753 \\
        TCN-RNN-ST & 15 & 3 & 32 & 1 & 32 & 0.88033 \\
        BiLSTM & - & - & - & 2 & 32 & 0.86577 \\
        TCN-ST & 11 & 4 & 32 & - & - & 0.80395 \\
        GRU & - & - & - & 4 & 16 & 0.63382 \\
        LSTM & - & - & - & 4 & 16 & 0.57495 \\
        \hline
        \end{tabular}\\
        F1-scores sorted from high to low. A dash ``-" indicates that the parameter is not applicable for that particular model. 
    \end{table*}

\begin{figure}[ht]
        \centering
          \includegraphics[trim={3.1cm 8.2cm 2.4cm 8.4cm},clip,width=\linewidth]{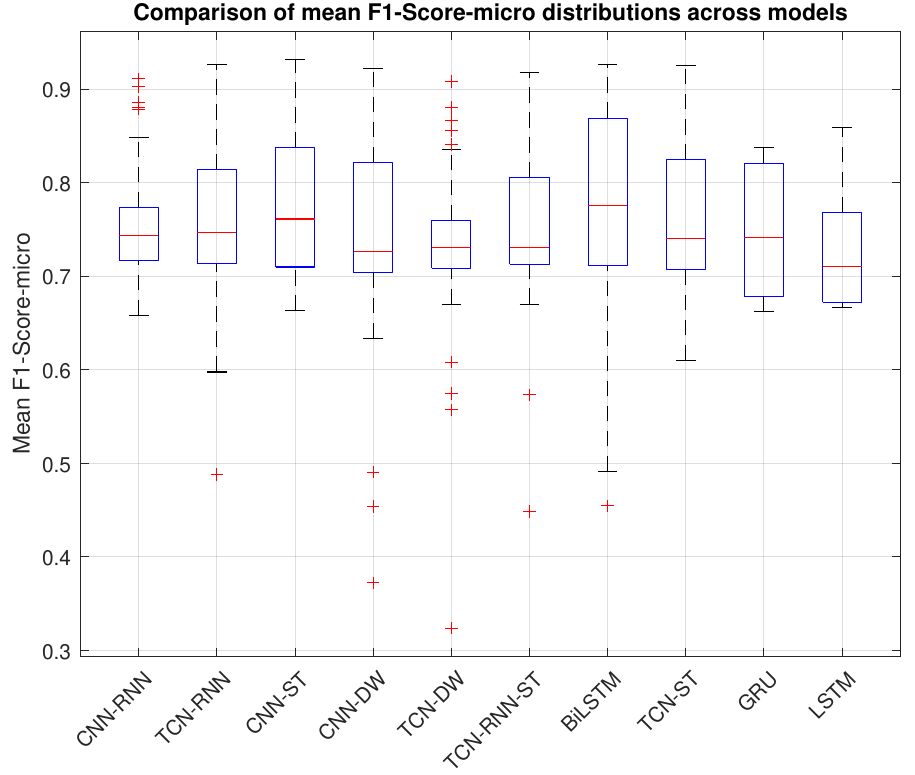}
          \caption{Box plot comparison of F1-Score-micro distributions across different deep learning models for EEG blink detection. CNN-RNN shows the highest median performance, while LSTM and CNN-DW exhibit the lowest performance among the tested architectures.}
          \label{fig:f1_micro_boxplot}
    \end{figure}

    \begin{figure}[ht]
        \centering
          \includegraphics[trim={3.1cm 8.2cm 2.4cm 8.4cm},clip,width=\linewidth]{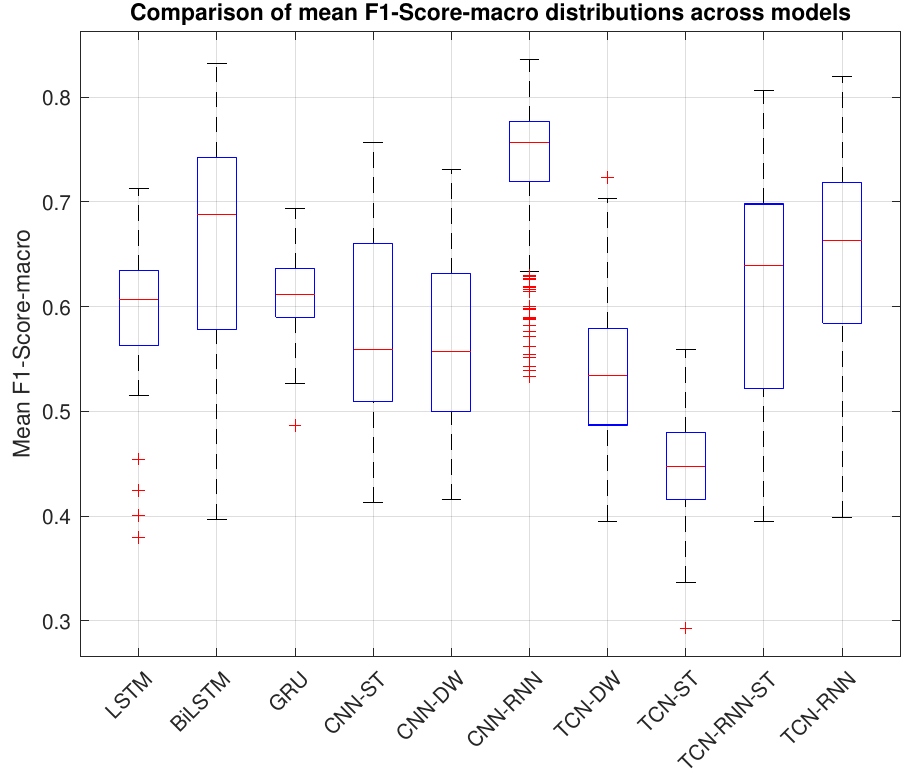}
          \caption{Box plot comparison of F1-Score-macro distributions across different deep learning models for EEG blink detection. CNN-RNN demonstrates the most consistent performance with minimal variance, while simpler RNN models show greater variability in results.}
          \label{fig:f1_macro_boxplot}
    \end{figure}
\section{Discussion}
    
    \subsubsection{Accuracy and Robustness}  
    The CNN-RNN model using BiLSTM consistently outperformed other models. This can be attributed to its ability to process data in both forward and backward directions, capturing more temporal dependencies and patterns in EEG signals. The superior performance across different channel configurations suggests that the BiLSTM model effectively utilizes the additional spatial information provided by multiple channels.
    
    \subsection{Model Performance Analysis}
 
    \subsubsection{Impact of EEG Channels}
    The number of EEG channels had a significant impact on the performance of the model. Models trained with three and five channels generally performed better than those trained with a single channel. However, it is important to note that during the grid search process, the prediction results for five-channel inputs were lower than those for one- and three-channel inputs. This is a rather anomalous phenomenon.
    Nevertheless, during the testing phase, the results for the multichannel inputs were still slightly better than those for the one- and three-channel inputs. This discrepancy could be due to the relatively small dataset used for the hyperparameter search.
    
    The improvement in using more channels can be attributed to the additional spatial information provided by multiple channels, which helps to distinguish blinks from other artifacts. However, the increased computational complexity and resource requirements associated with multi-channel models should be considered, especially for real-time applications.

    \subsubsection{Healthy Subjects vs. PD Patients}
    The models demonstrated robust performance in both healthy subjects and patients with Parkinson's disease. However, there were slight variations in accuracy, which could be due to the presence of tremors and other artifacts in the EEG signals of patients with PD. Despite these challenges, the models maintained high accuracy, indicating their potential for reliable blink detection in diverse populations.

    \subsubsection{Depthwise Separable Convolution}
    We have tested depthwise connection in CNN usage models. For fair comparison, the depthwise connection and standard versions are designed to have the same number of parameters. It does not show significant differences in accuracy except for the TCN-RNN model; the depthwise connection version has better results. Also, among all CNN usage models, under the same conditions, the training of the depthwise connection models is faster than the standard models. The reason may lie in the computational complexity of the depthwise connection being significantly lower than that of standard. This means that less computation is required for each forward and back pass. In addition, fewer memory accesses are generally required.

\newlength{\heightA}
\newlength{\heightB}

\setlength{\heightA}{29.7cm - 9.82cm - 8.34cm}
\setlength{\heightB}{29.7cm - 11.07cm - 10.5cm}

\newlength{\minHeight}
\setlength{\minHeight}{\minof{\heightA}{\heightB}}

\begin{figure*}[ht]
    \centering
    \begin{subfigure}[t]{0.14\textwidth}
    \includegraphics[trim={9.25cm 9.82cm 7.65cm 8.34cm},clip,width=\linewidth]{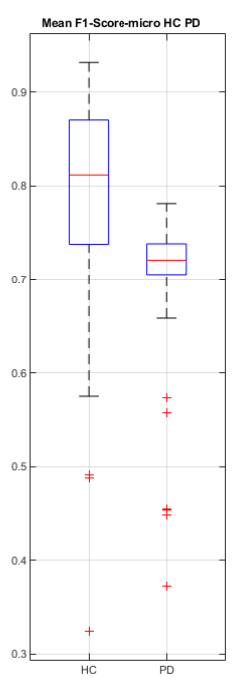}
    \caption{\label{fig:} }
    \end{subfigure}
    ~
    \begin{subfigure}[t]{0.84\textwidth}
        \centering
    \includegraphics[trim={3.45cm 12.1cm 3cm 10.5cm},clip,width=\linewidth]{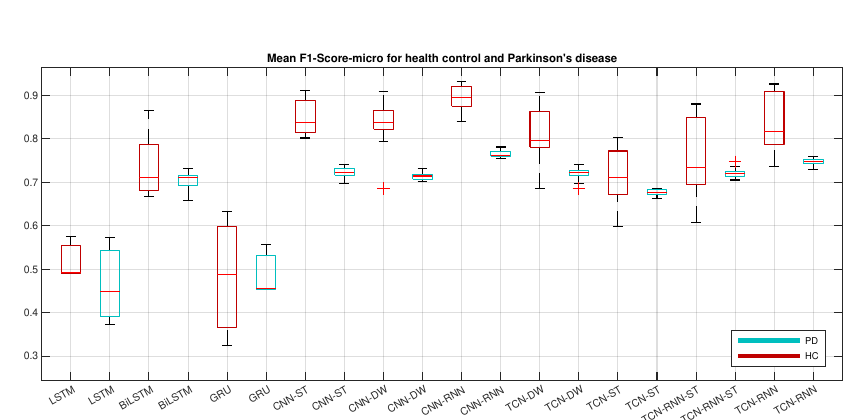}
    \caption{\label{fig:} }
    \end{subfigure}
\end{figure*}

    \subsection{Practical Implications}

    \subsubsection{Limitations and Challenges}
    Several limitations were encountered in this study. Although robust, the data set used may not cover all possible variations in EEG signals. Additionally, the preprocessing steps, while minimal, might still affect the generalizability of the models to raw, real-world data. The computational complexity of models, particularly those that use multiple channels, also poses a challenge for real-time applications and wearable devices.
    
    \subsection{Future Work}
    
    \subsubsection{Models Improvements}
    Future work should focus on improving the efficiency and generalizability of the models. This could include exploring more advanced architectures, such as attention mechanisms, or optimizing hyperparameters through more extensive searches. Incorporating domain adaptation techniques could also help the models generalize better to different datasets and recording conditions.

\section{Conclusion}
    The apparent difference in test results between Parkinson's disease patients and healthy controls is most likely due to the tremor that Parkinson's disease patients exhibit, and the presence of this tremor makes the blink signal unclear. In general, oscillation during tremors is believed to appear in the centroparietal region rather than in the forehead area. The differences reflected in the test indicate the possibility of further distinguishing Parkinson's syndrome by extracting blink signal features through forehead electrodes.

    
    This study presented several deep learning models for blink detection in EEG signals. The CNN-RNN model demonstrated the highest accuracy. Although the models showed robust performance across different channel configurations and populations, future work should address their limitations and explore extended applications. The findings highlight the potential of deep learning for accurate and efficient physiological signal processing segmentation.


\section{Appendix}

\subsection{CNN-DW Results}

\begin{figure*}[t!]
    \centering
    \begin{subfigure}[t]{0.3\textwidth}
        \centering
        \includegraphics[width=1\linewidth]{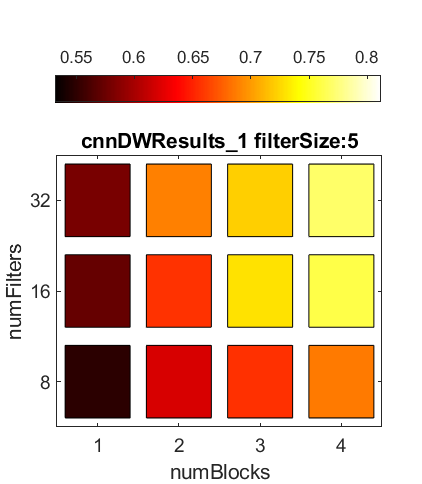}
        \caption{Filter Size: 5}
    \end{subfigure}%
    \begin{subfigure}[t]{0.3\textwidth}
        \centering
        \includegraphics[width=1\linewidth]{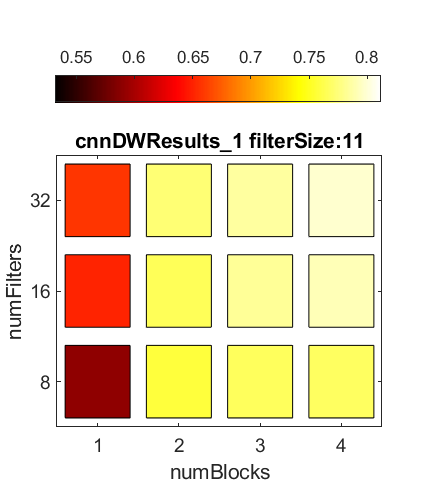}
        \caption{Filter Size: 11}
    \end{subfigure}
    \centering
    \begin{subfigure}[t]{0.3\textwidth}
        \centering
        \includegraphics[width=1\linewidth]{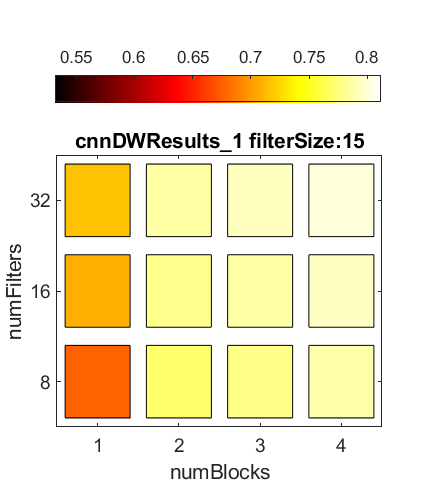}
        \caption{Filter Size: 15}
    \end{subfigure}\\ 

    \centering
    \begin{subfigure}[t]{0.3\textwidth}
        \centering
        \includegraphics[width=1\linewidth, trim=0 0 0 2cm, clip]{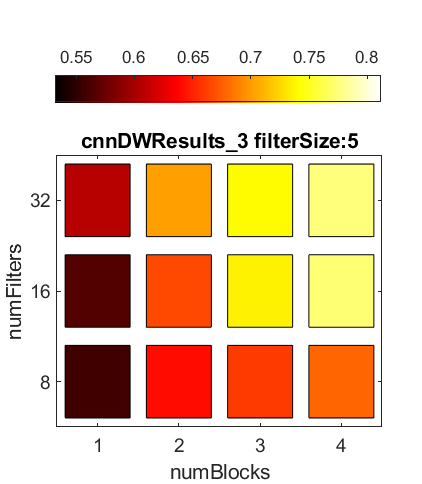}
        \caption{Filter Size: 5}
    \end{subfigure}%
    \begin{subfigure}[t]{0.3\textwidth}
        \centering
        \includegraphics[width=1\linewidth, trim=0 0 0 2cm, clip]{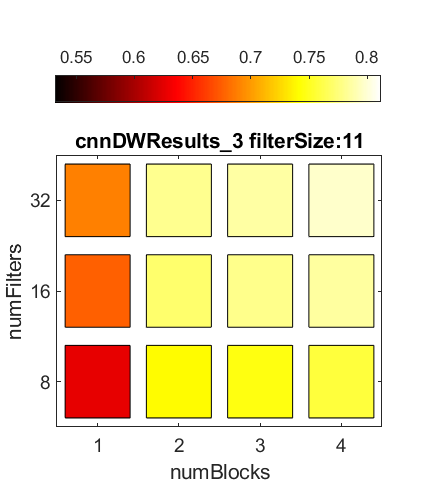}
        \caption{Filter Size: 11}
    \end{subfigure}
    \centering
    \begin{subfigure}[t]{0.3\textwidth}
        \centering
        \includegraphics[width=1\linewidth, trim=0 0 0 2cm, clip]{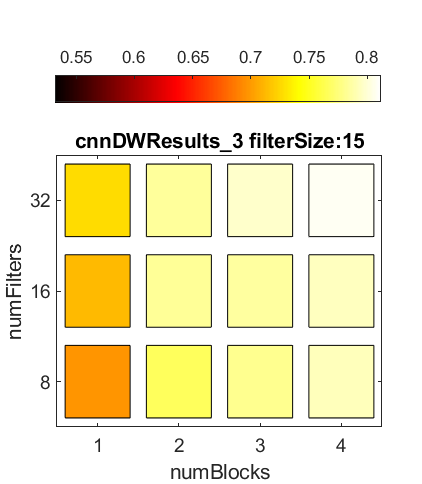}
        \caption{Filter Size: 15}
    \end{subfigure}\\ 

    \centering
    \begin{subfigure}[t]{0.3\textwidth}
        \centering
        \includegraphics[width=1\linewidth, trim=0 0 0 2cm, clip]{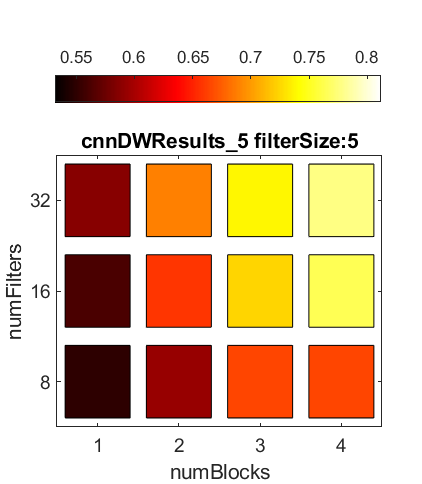}
        \caption{Filter Size: 5}
    \end{subfigure}%
    \begin{subfigure}[t]{0.3\textwidth}
        \centering
        \includegraphics[width=1\linewidth, trim=0 0 0 2cm, clip]{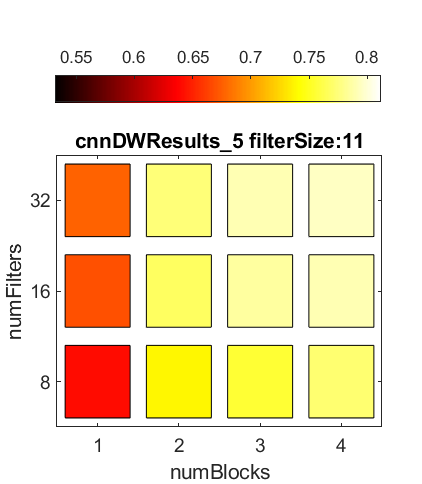}
        \caption{Filter Size: 11}
    \end{subfigure}
    \centering
    \begin{subfigure}[t]{0.3\textwidth}
        \centering
        \includegraphics[width=1\linewidth, trim=0 0 0 2cm, clip]{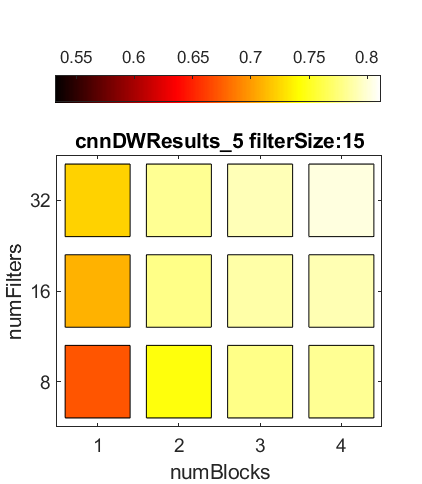}
        \caption{Filter Size: 15}
    \end{subfigure}\\ 
    
    \caption{Hyperparameter search for CNN over filter size, number of filters and number of blocks.}
\end{figure*}

\end{document}